\documentclass[review]{elsarticle}

\usepackage{enumerate}
\usepackage{amsmath}
\usepackage{amssymb}
\usepackage{booktabs}
\usepackage{tabularx}
\usepackage{graphicx}
\usepackage{subcaption}
\usepackage[margin=1in]{geometry}

\usepackage[normalem]{ulem}
\usepackage{xcolor} 

\usepackage{url}
\begin{document}
\date{}
\begin{frontmatter}

\title{A Structural Feature-Based Approach for Comprehensive  Graph Classification}

\author[affiliation_1,affiliation_2]{Saiful Islam\corref{mycorrespondingauthor1}}
\cortext[mycorrespondingauthor1]{Corresponding author}
\ead{saifulis@buffalo.edu}
\author[affiliation_3]{Md. Nahid Hasan}
\author[affiliation_4]{Pitambar Khanra}

\address[affiliation_1]{Institute for Artificial Intelligence and Data Science,University at Buffalo, State University of New York at Buffalo, NY 14260, USA}
\address[affiliation_2]{Department of Mathematics, University of Dhaka, Dhaka 1000, Bangladesh}
\address[affiliation_3]{Department of Mathematics, Bangladesh University of Engineering and Technology, Dhaka 1000, Bangladesh}
\address[affiliation_4]{Jacobs School of Medicine \& Biomedical Sciences, University at Buffalo, State University of New York at Buffalo, NY 14203, USA}


\begin{abstract}
The increasing prevalence of graph-structured data across various domains has intensified greater interest in graph classification tasks. While numerous sophisticated graph learning methods have emerged, their complexity often hinders practical implementation. In this article, we address this challenge by proposing a method that constructs feature vectors based on fundamental graph structural properties. We demonstrate that these features, despite their simplicity, are powerful enough to capture the intrinsic characteristics of graphs within the same class. We explore the efficacy of our approach using three distinct machine learning methods, highlighting how our feature-based classification leverages the inherent structural similarities of graphs within the same class to achieve accurate classification. A key advantage of our approach is its simplicity, which makes it accessible and adaptable to a broad range of applications, including social network analysis, bioinformatics, and cybersecurity. Furthermore, we conduct extensive experiments to validate the performance of our method, showing that it not only reveals a competitive performance but in some cases surpasses the accuracy of more complex, state-of-the-art techniques. Our findings suggest that a focus on fundamental graph features can provide a robust and efficient alternative for graph classification, offering significant potential for both research and practical applications.
\end{abstract}


\end{frontmatter}


\section{Introduction}\label{sec:introduction}
A graph, often referred to as a network, is a collection of nodes (vertices) and edges (links between vertices)~\cite{wasserman1994social, barabasi2004network,  newman2018networks, barabasi2012network}. We use the term graph and network alternatively. The exploration of network is omnipresent, finding applications in unraveling complex phenomena across diverse interconnected systems surrounding us such as ecosystems, social systems~\cite{dorogovtsev2003evolution, borgatti2009network, morales2017global, newman2018networks, shahal2020synchronization, hasan2021ivesr}, biological~\cite{dorogovtsev2003evolution, bassett2006small, bick2020understanding} and physical systems~\cite{newman2003structure, boccaletti2006complex, newman2011structure, estrada2012structure}, and more. Graph classification, a crucial subtask in many application areas of network sciences and graph mining~\cite{riesen2009graph, cai2018comprehensive}, plays an important role in diverse fields such as biological, physical, chemical, social science, and technologies~\cite{lebichot2016graph, ma2018graph,zhang2020deep, xia2021graph, dara2022machine, xie2022active, ju2022kgnn}. In biological sciences, for example, graph classification is employed to classify different types of tissues and cells, aiding in the discovery of new drugs and improving therapeutic measures~\cite{tran2021deep, gaudelet2021utilizing, ahmedt2021graph}. Similarly, in chemical sciences, graph classification helps determine molecular properties, such as solubility~\cite{dara2022machine}. In neuroscience, this technique is used to achieve accurate diagnoses of conditions like autism spectrum disorder~\cite{shao2023heterogeneous}. Furthermore, in the field of network science, graph classification is instrumental in precisely detecting rumors in social networks~\cite{monti2019fake}. Beyond these fields, graph classification has important applications in text classification, natural language processing~\cite{lai2015recurrent, huang2019text, molokwu2020node, wang2023graph, wu2023graph}. For instance, it can classify various types of social networks, such as determining movie genres based on actors' and actresses' ego networks or identifying fraudulent transaction within financial networks~\cite{riesen2009graph, lebichot2016graph, xia2021graph}.

Graph classification involves assigning graphs in a database to different classes. When graph labels are available, classification tasks are performed in a supervised manner, whereas, in the absence of labels, graphs can be clustered into various groups using unsupervised way~\cite{rehman2012graph}. Various approaches have been employed for graph classification, including matrix factorization, graph neural network (GNN), kernel-based methods, and graph embedding techniques~\cite{zhu2007combining, perozzi2014deepwalk, scarselli2008graph, cai2018comprehensive, cai2018comprehensive, vishwanathan2010graph, grover2016node2vec, goyal2018graph, wu2020comprehensive, errica2019fair, ju2022kgnn, kriege2020survey}. GNNs, for example, update a node's features by iteratively aggregating and pooling its neighbor's features. Different GNNs use various pooling methods, such as sum pooling, hierarchical pooling, and differentiable pooling~\cite{kipf2016semi, hamilton2017inductive, velivckovic2017graph, xu2018powerful, cangea2018towards, ying2018hierarchical}, to extract a feature vector representing the entire network.  The embedded network is then passed through fully connected layers, followed by a softmax or sigmoid layer, to predict graph labels. Graph kernel-based methods, on the other hand, decompose the entire graph into subgraphs and use these to measure the similarity between graphs by computing the inner products~\cite{gartner2003graph, shervashidze2009efficient, shervashidze2011weisfeiler, kriege2020survey}. Several well-known node embedding algorithms are used for graph classification by embedding nodes into Euclidean space~\cite{grover2016node2vec, grohe2020word2vec}. After obtaining the vector representation of nodes, aggregating these vectors yields the representation of the entire graph. Additionally, some systems employ hand-crafted graph structural attributes as features to represent the graph~\cite{chen2019powerful, ma2020graph, huynh2022graph, sislam2024}. Various classification approaches are then used for the classification tasks~\cite{zhu2007combining,scarselli2008graph, vishwanathan2010graph, perozzi2014deepwalk, grover2016node2vec, errica2019fair, goyal2018graph, ju2022kgnn}.

However, implementing methods such as GNNs, matrix factorization, kernel-based approaches, and graph embedding involves considerable subtleties and interpretation challenges~\cite{riesen2009graph, baldini2020complexity}. Many existing methods encounter challenges from high computational and space costs due to reliance on high dimensional feature spaces and requiremet of large amount of data, which negatively impact the classification performance~\cite{cai2018comprehensive,zhang2018end}. Matrix factorization algorithms, notwithstanding their effectiveness, demand significant time and space resources~\cite{cai2018comprehensive}, and graph kernel approaches struggle with dimensionality issues due to the dependent nature of the substructures~\cite{cai2018comprehensive}. GNNs have achieved remarkable success, however they are frequently data-hungry, requiring a significant quantity of data for training and can be limited by complex feature spaces~\cite{riesen2009graph, hao2020asgn, liu2024frequency}. Several studies have focused on exploring graph structural features for classification~\cite{li2011graph, bonner2016deep, xuan2019subgraph, ma2020graph, huynh2022graph}, but these often involve computationally demanding procedures. Some authors have used neural network approaches for graph embedding, which increases computational overhead ~\cite{bonner2016deep, xuan2019subgraph, ma2020graph, huynh2022graph}.

Given these challenges, there is a compelling need for a simple yet effective strategy for graph classification. In this study, we propose a straightforward yet effective ML-based approach for graph classification, leveraging nine easily computable key structural properties: number of nodes, number of edges, average degree, diameter, closeness centrality, betweenness centrality, clustering coefficient, spectral radius, and trace of the Laplacian matrix. 
These properties are considered effective because they capture a wide range of structural aspects of a network. For instance, the number of nodes and edges provides fundamental scale and density metrics~\cite{watts1998collective, barabasi1999emergence, newman2018networks}, while the average degree offers insights into connectivity levels. The diameter measures the longest shortest path between any two nodes, reflecting communication efficiency~\cite{barabasi1999emergence}. Average closeness centrality assesses how easily nodes can reach others, and betweenness centrality identifies influential intermediaries~\cite{freeman2002centrality}. The clustering coefficient quantifies local groupings or clusters within the network~\cite{watts1998collective}. Additionally, the spectral radius and trace of the Laplacian matrix provide information on the network's overall connectivity and robustness~\cite{chung1997spectral}. By incorporating these varied aspects into a feature vector, our approach achieves a nuanced and comprehensive understanding of network structures, which enhances the accuracy of graph classification by effectively distinguishing between different types of networks, such as dense versus sparse or well-connected versus fragmented ones and so on. Our method aims to bridge the gap between simplicity and performance, offering a viable alternative to more complex techniques. We evaluate our approach using ten different benchmark datasets namely COLLAB~\cite{yanardag2015deep, KKMMN2016}, IMDB-BINARY (IMDB-B), IMDB-MULTI (IMDB-M)~\cite{yanardag2015deep, ivanov2019understanding, Morris+2020}, REDIT-BINARY (RDT-B), REDIT-MULTY5K(RDT-M5K)~\cite{yanardag2015deep}, PROTEINS~\cite{dobson2003distinguishing, borgwardt2005protein}, MUTAG~\cite{yanardag2015deep, ivanov2019understanding, Morris+2020}, NCI1, NCI109~\cite{wale2008comparison, wang2019research, Morris+2020}, and PTC~\cite{helma2001predictive, kriege2012subgraph}. 
For the classification task, we employ three different methods: k-Nearest Neighbors (k-NN)~\cite{fix1989discriminatory, bishop2006pattern},  Support Vector Machine (SVM)~\cite{hearst1998support, bishop2006pattern} and Random Forest~\cite{breiman2001random, biau2012analysis} classifier. We compare the classification performance of these methods and benchmark our approach against other state-of-the-art classification techniques.

\section{Methods}
\label{sec:methods} 
\begin{figure*}[htp]
\centering
\includegraphics[width=1.00\linewidth]{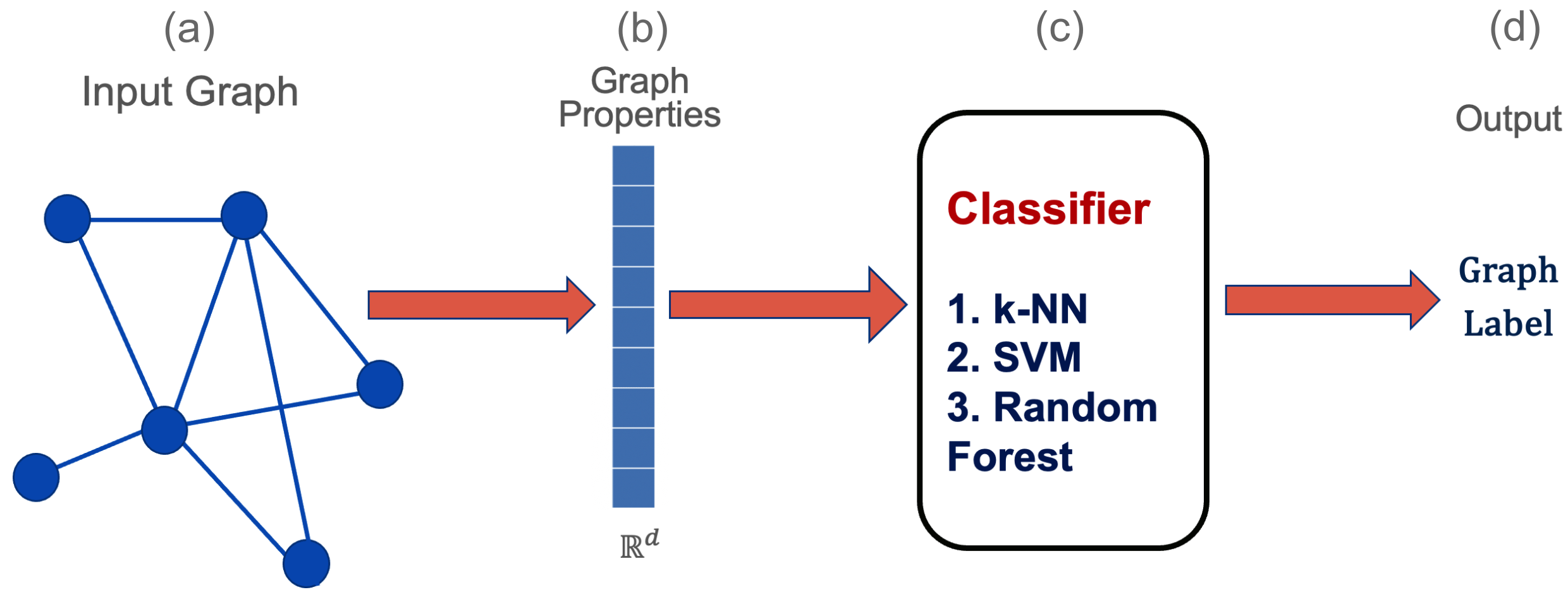}
\caption{\textbf{Schematic diagram of the proposed classification approach.} (a) Input the graph. (b) Calculated $d=9$ dimensional feature vector using nine graph properties. (c) The classifiers to predict the label of the given graph. (d) The output graph label.}\label{fig:methods}
\end{figure*}

In this section, we briefly describe the datasets, graph properties used in the feature vectors, and classification methods used in this study. Figure~\ref{fig:methods} shows a high-level overview of our methodology, where the initial input is a graph. We next compute a $d=9$ dimensional feature vector for each graph. The classifier then uses this feature vector to predict the label associated with the provided graph.

\subsection{Datasets} 
We use ten well-known benchmark datasets. Five of them are social network datasets: COLLAB~\cite{yanardag2015deep, KKMMN2016}, IMDB-B, IMDB-M~\cite{yanardag2015deep, ivanov2019understanding, Morris+2020}, RDT-B, and RDT-M5K~\cite{yanardag2015deep}. The remaining six are biological network datasets: PROTEINS~\cite{dobson2003distinguishing, borgwardt2005protein}, MUTAG~\cite{debnath1991structure, kriege2012subgraph}, NCI1, NCI109~\cite{wale2008comparison, wang2019research, Morris+2020}, and PTC~\cite{helma2001predictive, kriege2012subgraph}
Various studies in graph learning methods use these datasets to assess the performance of their methods~\cite{errica2019fair, hu2020open, Morris+2020, wu2020comprehensive}.

\subsubsection{Social network}
Our first  social network dataset is a scientific collaboration dataset, namely, COLLAB. Graphs in this dataset are ego networks of the researchers from one of the three physics fields: Condensed Matter Physics, High Energy Physics, and Astro Physics. Therefore, these three fields indicate the class of graphs. In the ego network, each node represents a researcher. Two nodes are connected with an edge if the corresponding researchers have a collaboration work \cite{yanardag2015deep, KKMMN2016, Morris+2020}.

Two of the five social network datasets are movie collaboration networks which are IMDB-B dataset and IMDB-M dataset. The IMDB-B dataset consists of the ego networks of movie actors and actresses. Actors or actresses are represented by nodes, and an edge between any two nodes shows that these two actors or actresses have played roles in the same movie. The networks are categorized into two classes: Romance and Action genres ~\cite{yanardag2015deep, ivanov2019understanding, Morris+2020}. The IMDB-M dataset, which consists of the similar graphs to the IMDB-B datasets. That is, nodes represent actors and actresses and edge indicates their collaboration in the same types of movies. However, the edges are taken from three distinct genres: Sci-Fi, comedy and romance \cite{yanardag2015deep, ivanov2019understanding, Morris+2020}.

We also use two other social network datasets from online discussions on Reddit which are RDT-B dataset and RDT-M5K dataset. In the RDT-B dataset, every node represents a user in the network and every edge portrays the responses of any individual user to other’s comments \cite{yanardag2015deep}. The graphs are labeled based on whether they are part of a discussion community or a question-and-answer community. The RDT-M5K dataset is a collection of five different subreddits, namely, worldnews, videos, AdviceAnimals, aww and mildlyinterestng \cite{ yanardag2015deep}. The graphs are labeled based on the corresponding subreddits.

\subsubsection{Biological network}
The PROTEINS dataset, each graph represents a protein structure. In the graph, the nodes are the amino acids and there is an edge between two amino acids if the distance between two amino acids is less than 0.6nm ~\cite{dobson2003distinguishing, borgwardt2005protein}. These graphs are classified as enzymes or non-enzymes \cite{dobson2003distinguishing, borgwardt2005protein, Morris+2020}.

The MUTAG dataset is a collection of chemical compounds with two classes. The classes are the heteroaromatic and mutagenic aromatic nitro compounds that were examined for mutagenicity on Salmonella typhimurium. In each chemical compound graph, the nodes are the atoms in the compound and the edges are chemical bonds between two atoms \cite{debnath1991structure, kriege2012subgraph}.

We use two cheminformatics datasets published by the National Cancer Institute (NCI), NCI1 and NCI109 datasets. These datasets are used to classify chemical compounds. These two benchmark data sets are mainly used for forecasting anti-tumor activity \cite{wale2008comparison, wang2019research, Morris+2020}. Here, each node represents an atom in the molecules, while the edges connecting the nodes indicate atomic bonds. The graphs are classified as positive or negative to cell lung cancer.

The Predictive Toxicology Challenge (PTC) dataset is known as the small molecule dataset. This dataset is a part of the TUDataset collection \cite{ wang2019research}. PTC-MR comprises graphs from two classes that show the impact of chemical compound carcinogenicity on male rats \cite{ helma2001predictive, kriege2012subgraph}. Similar to the NCI1 and NCI109, nodes represent atoms and edges represent chemical bonds.

The statistical summary of these datasets are shown in Tables~\ref{tab:datasets-soci}, and~\ref{tab:datasets-bio}. The first row indicates the number of graphs in each dataset, the second  and fifth rows are the number of average nodes and edges of graphs, and the third and the fourth rows represent the maximum and minimum number of nodes in a graph in the corresponding dataset.  The sixth row indicates the number of classes of graphs.

\begin{table*}
\centering
\caption{\textbf{Statistical summary of the social networks in the benchmark datasets.} The rows indicating ``Avg. nodes", ``Max nodes", ``Min nodes", and ``Avg. edges" denote the average number of nodes, maximum number of nodes, minimum number of nodes, and average number of edges of the graphs respectively, in the corresponding dataset.} 
\label{tab:datasets-soci}
\begin{tabular}{lrrrrr}
\toprule
Dataset & COLLAB &  IMDB-B & IMDB-M & RDT-B & RDT-5KM \\
\midrule
Number of graphs &  5,000 & 1,000 & 1,500 & 2,000 & 4,999 \\
Avg. nodes  & 74.49  & 19.77  & 13.00  & 429.63 & 508.52 \\
Max nodes & 492 & 136 & 89 & 3,782 &  3,648\\
Min nodes  & 32 & 12 & 7 & 6 & 22\\
Avg. edges   & 2,457.78 & 96.53 & 65.94 & 497.75 & 594.87 \\
Max edges & 40,119 & 1,249 & 1,467 & 4,071 &  4,783\\
Min edges & 60 & 26 & 12 & 4 & 21\\
Number of classes    & 3   &  2 & 3 & 2 & 5 \\
\bottomrule
\end{tabular}
\end{table*}

\begin{table*}
\centering
\caption{\textbf{Statistical summary of the biological networks in the benchmark datasets.} The rows indicating ``Avg. nodes", ``Max nodes", ``Min nodes", and ``Avg. edges" denote the average number of nodes, maximum number of nodes, minimum number of nodes, and average number of edges of the graphs respectively, in the corresponding dataset.} 
\label{tab:datasets-bio}
\begin{tabular}{lrrrrr}
\toprule
Dataset & PROTEINS & MUTAG  & NCI1 & NCI109  & PTC   \\
\midrule
Number of graphs & 1,113   & 188  & 4,110 & 4,127 & 344 \\
Avg. nodes & 39.06   & 17.93   & 29.87  & 29.68  & 14.29  \\
Max nodes &  620 &  28 & 111 & 111 &  64 \\
Min nodes & 4 & 10 & 3 & 4 & 2 \\
Avg. edges & 72.82  & 19.79    & 32.30 &  32.13 & 14.69 \\
Max edges &  1,049 &  33 & 119 & 119 &  71\\
Min edges & 5 & 10 & 2 & 3 & 1 \\
Number of classes & 2  & 2   & 2   &  2 & 2 \\
\bottomrule
\end{tabular}
\end{table*}

\subsection{Network preliminaries}
We construct a nine dimensional feature vector, representing each network using the nine network structural properties to classify the networks. The nine structural properties are respectively, (1) number of nodes, (2) number of edges, (3) average degree of the network, (4) diameter, (5) average closeness centrality, (6) average betweenness centrality, (7) average clustering coefficient, (8) spectral radius of the Laplacian matrix, and (9) the trace of the Laplacian matrix corresponding to the network. The definitions of the nine properties and other related properties are as follows.\\

\begin{enumerate}
    \item [(1)] The first property that we use as a feature of a network is the number of nodes in that network. We denote the number of nodes as $n$. 

   \item []  A network with $n$ nodes can be represented by a $n \times n$ dimensional adjacency matrix given by $A=[A_{ij}]$, where $A_{ij}$ denotes the $(i, j)$th entry of the adjacency matrix and is defined as~\cite{newman2018networks}.
\begin{equation}
    A_{ij}  = \begin{cases}
        1, & \text{ if node } i \text{ and node } j \text{ is connected}\\
        0, & \text{otherwise}.
    \end{cases}
\end{equation}
We use the properties related to the adjacency matrix and/or can be derived from the adjacency matrix.
\item [(2)] The second property that we include in the  feature vector is the number of edges of the network. For a  given network with the adjacency matrix $A$, the number of edges is given by~\cite{newman2018networks}. 
\begin{equation}
    m = \frac{1}{2}\sum_{i=1}^n \sum_{j =1}^n A_{ij}.
\end{equation}

\item[] The degree of a node is determined by the number of neighbors of that node in the network. The degree of the node $i$ is denoted by $k_i$ and defined by~\cite{newman2018networks},
\begin{equation}
    k_i = \sum_{j=1}^n A_{ij}.
\end{equation}
\item[(3)]  As a third property, we use the average degree of the network which is given by~\cite{newman2018networks},
\begin{equation}
    \langle k \rangle = \frac{1}{n} \sum_{i=1}^n k_i.
\end{equation}

\item [(4)] The diameter of a network is defined as the largest value of all the shortest paths in the network~\cite{newman2018networks}. In the case of a disconnected network, we use the diameter of the largest component since the length of the shortest path between two nodes in two different components is undefined. 

\item [(5)]  We also use the average closeness centrality of the graph. The closeness centrality of a node $i$ is defined by~\cite{sabidussi1966centrality, newman2018networks}.
\begin{equation}
    H_i = \frac{n-1}{ \sum_j^n d_{ij}},
\end{equation}
where $d_{ij}$ is the shortest distance between node $i$ and node $j$.

\item [(6)] Similarly, we compute the average betweenness centrality of the graph. The betweenness centrality of a node $i$ is defined as follows~\cite{sabidussi1966centrality, newman2018networks}.
\begin{equation}
    B_i = \frac{\sum_{st} n_{st}^i}{ \sum_{st} n_{st}},
\end{equation}
where $n_{st}$ is the number of shortest paths between node $s$ and $t$, and $n_{st}^i$ is the number of shortest paths between node $s$ and $t$ passes through the node $i$. 

\item [(7)]  Average clustering coefficient. The local clustering coefficient of node $i$ in the network is defined by~\cite{newman2018networks}.
\begin{equation}
    c_i =\frac{\text{Number of pairs of neighbors of node } i \text{ that are connected}}{ \text{Number of pairs of neighbors of node }i}.
\end{equation}
We use the average clustering coefficients of the network~\cite{newman2018networks}.

\begin{equation}
    C = \frac{1}{n} \sum_i c_i.
\end{equation}

\item[(8)] In our study, we  also use the spectral radius of the Laplacian matrix $L$.
\item [] The Laplacian matrix is one of the important representations of a network which can be represented as an $n \times n$ matrix with elements $L_{ij}$ defined by~\cite{newman2018networks}.  
\begin{equation}
L_{ij}= \begin{cases}
  k_i,  &  \text{if } i =j\\
  -1, &  \text{ if node } i \text{ and } j \text{ is connected}\\
  0, & \text{ otherwise}.
\end{cases}
\end{equation}
In matrix form it can be written as $L=D-A$, where $D$ is the diagonal degree matrix of the network which is given by,
\begin{equation}
    D = \begin{pmatrix}
        k_1 & 0 & 0 & \dots & 0\\
        0 & k_2 & 0 & \dots & 0\\
        \vdots &  & \ddots &  & \vdots\\
        0 & 0 & 0 & \dots & k_n\\
    \end{pmatrix}.
\end{equation}
and $A$ is the corresponding adjacency matrix.
If $\lambda_1, \lambda_2, ..., \lambda_n$ are the eigenvalues of $L$, then the spectral radius is given by 
\begin{equation}
    \rho(L) = \text{max}\{ |\lambda_1|, |\lambda_2|, ..., |\lambda_n|\}.
\end{equation}
\item[(9)] We also use the trace of the matrix $L$ which is given by 
\begin{equation}
    \text{Tr}(L) = \lambda_1+\lambda_2+...+\lambda_n.
\end{equation}

\end{enumerate}
To compute the feature vector for each of the networks, we use the \texttt{Python} library \texttt{NetworkX}~\cite{SciPyProceedings_11}, and \texttt{pandas}~\cite{reback2020pandas} for preparing the feature vectors as Dataframe. For visualization, we employ the \texttt{seaborn}~\cite{Waskom2021}  library.

\subsection{Classification methods} \label{classifier}
There are many effective classification methods used for classification tasks. In this study,  we only use the three popular but simple classification methods, (i) k-nearest neighbors algorithm (k-NN), (ii) support vector machine (SVM), and (iii) Random Forest (RF) method. In this section, we briefly describe these three methods.

\subsubsection{k-Nearest Neighbors}
The k-NN is a non-parametric supervised machine learning method which is used for both regression and classification tasks~\cite{fix1989discriminatory, bishop2006pattern}. This is one of the simplest machine learning methods widely used since it has been introduced~\cite{fix1989discriminatory, bishop2006pattern}. As our aim is the graph classification, we use k-NN as a classifier. For this approach there is no explicit training phase. It uses all the data points in the training set for classifying a selected data point $\overline{x}^*$. To classify the selected  data point $\overline{x}^*$,  the method follows the below steps~\cite{bishop2006pattern}.
\begin{enumerate}
    \item Choose a user defined value of $k$, the number of nearest neighbors.
    \item Select k-nearest neighbors of the data point $\overline{x}^*$  from all the data points $\overline{x}$, based on the user given distance function.  To determine the k-nearest neighbors, we use the Euclidean distance given by,
    \begin{equation}
        d(\overline{x}, \overline{x}^*) = \sqrt{\sum_i^d (x_i-x_i^*)^2}.
    \end{equation}
    \item Determine the class label of the point $\overline{x}^*$ based on the majority vote of the k-nearest neighbors. 
\end{enumerate}
We utilize the \texttt{KNeighborsClassifier} in \textit{scikit-learn}  package in Python~\cite{pedregosa2011scikit}. 

\subsubsection{Support Vector Machine}
The support vector machine is another powerful supervised learning method for classifying data into different classes. It determines a hyperplane that separates the data points into classes. Indeed, it maximizes the margin between the hyperplane and the nearest data points from the hyperplane on both sides~\cite{hearst1998support, bishop2006pattern, hastie2009elements}. These nearest vectors to the hyperplane are referred to as support vectors. The hyperplane in defined by 
\begin{equation}
    y = \overline{w}^\top \overline{x}+b,
\end{equation}
where $\overline{x}$ is a feature vector, $\top$ denotes the transpose. The points are classified according to the sign of $y$ ~\cite{bishop2006pattern}. If there are only two classes in the datasets for one of them $y>1$ and for another one $y<-1$ then maximizing the marginal is equivalent to the following problem~\cite{hearst1998support,bishop2006pattern}.

\begin{eqnarray}
    & \text{Minimize } \frac{1}{2} ||\overline{w}||^2 \\
    & \text{Subject to:  }  \text{ Min } |y(\overline{w}^\top \overline{x}+b)| =1
\end{eqnarray}
The above description is the backbone of the SVM. The SVM uses kernel trick to transform feature space while the data points are not separable by using a simple hyperplane \cite{hearst1998support}. In this study,  we utilize the linear kernel and the built-in library \texttt{svm} in \textit{scikit-learn}~\cite{pedregosa2011scikit} to classify the graphs with the regularization parameter $C=10$.  

\subsubsection{Random Forest}
The Random Forest algorithm is an ensemble learning technique consisting of decision trees. It is widely used as a supervised approach and serves as a versatile tool for both classification and regression purposes in machine learning~\cite{breiman2001random, biau2012analysis}. A decision tree consists of nodes: root nodes, internal nodes, and terminal nodes~\cite{breiman1984classification, hastie2009elements}. The starting node is called the root node from where the internal nodes are descended. The nodes that facilitate data point segregation based on assigned features are known as internal nodes, while those that do not split further are referred to as leaf nodes or terminal nodes. Each internal node corresponds to a specific input feature, and each leaf node is labeled with a class~\cite{hastie2009elements}.   In the process of classification, a data point starts its journey from the root node and traverse the tree based on its feature value and arrives at a terminal node.  Finally, the data point is assigned the class associated with that leaf~\cite{hastie2009elements}. 

Random Forest introduces an ensemble approach by constructing multiple decision trees using bootstrapped training samples~\cite{breiman2001random, biau2012analysis}. Additionally, it leverages a subset of features, randomly sampled, as split candidates from the full set of features. To classify the input data point, the Random Forest uses the majority voting of all developed trees~\cite{breiman2001random, biau2012analysis}. We use the \texttt{RandomForestClassifier} class from \textit{scikit-learn}  package~\cite{pedregosa2011scikit} in Python with 200 trees in the forest, ``entropy" function to measure the quality of a split and maximum depth of trees is five.

\subsection{Computational complexity}
Generally, we can obtain the number of nodes and edges of a graph from the graph database. If these are not provided, we can calculate the number of nodes of a graph in time $O(n)$ and the number of edges in time $O(m)$. We can compute the average degree of a given graph in $O(n+m)$ time. Since most of the real world graphs are sparse, we can compute the Laplacian spectrum with complexity $O(n^2)$~\cite{meyer2023matrix}. The computational time for the clustering coefficient of a node is $O(n k_{max}^2)$, where $k_{max}$ is the highest degree of nodes in the graph. However, we compute the average clustering coefficient over all the nodes with the cost $O\left(\frac{m^2}{n}\right)$ using a better approximation for the maximum degree, $O(k_{max}^2) \approx \frac{2m^2}{n^2}$. The diameter of a sparse graph can be approximated  in $O(m\sqrt{n})$ time~\cite{roditty2013fast}. The computational complexity for the closeness centrality is $O(n^2+nm)$~\cite{duron2020heatmap} and for the betweenness centrality is $O(nm)$~\cite{brandes2001faster}.

\section{Results}\label{sec:results}
\begin{figure*}
\centering
\includegraphics[width=1.01\linewidth]{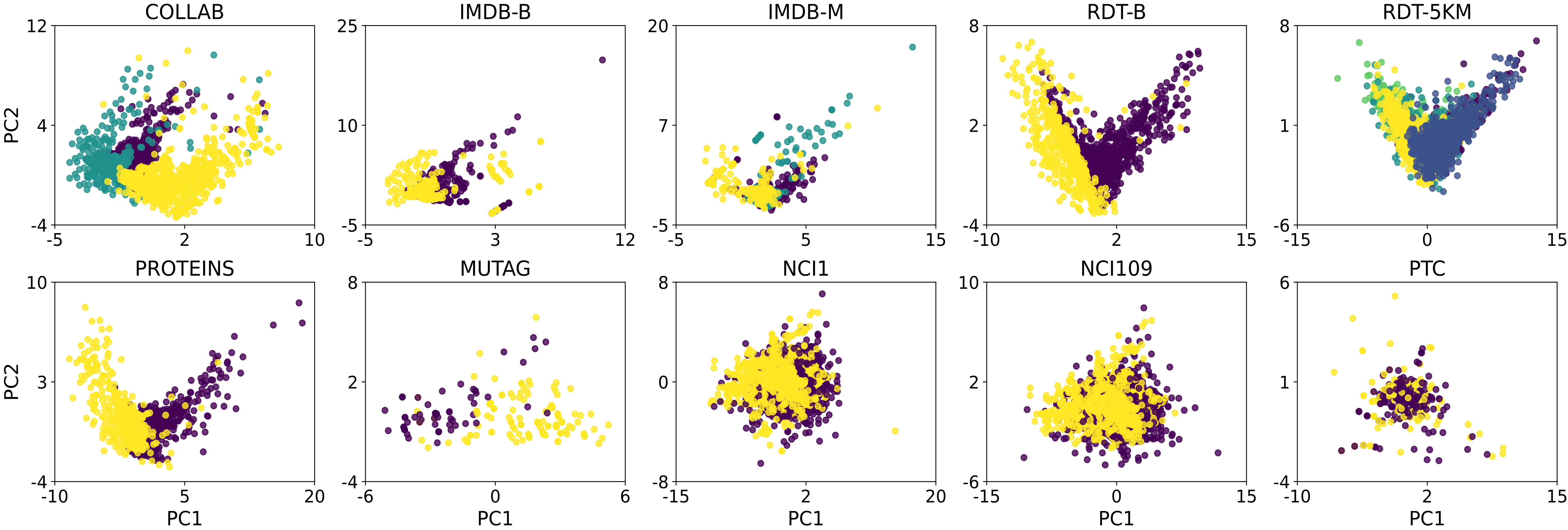}
\caption{Visualization of the two-dimensional embedding for the ten datasets. PC1: the first principal component, and PC2: the second principal component. Each circle represents a graph, and colors indicate the graph classes. }
\label{fig:pca}
\end{figure*}

We obtained a nine-dimensional feature vector for each graph. We then applied principal component analysis (PCA) to generate a two-dimensional embedding for each dataset. We visualize the resulting two-dimensional embeddings: the first  principal component, PC1, and the second principal component, PC2, for ten datasets in Fig.~\ref{fig:pca}. We observe that the graphs from different classes are reasonably separated in two dimensional embedding space. In certain datasets, such as COLLAB, and PTC, there are no clearly distinct boundaries between graph classes. However, in some cases like IMDB-B, RDT-B, PROTEIN, and MUTAG, the distinctions between graph classes are very clear. This visualization suggests that the classifiers can effectively differentiate between graph classes. Furthermore, in unsupervised learning scenarios where graph labels are unavailable, different clustering methods can potentially identify distinct graph groups.

\begin{figure*}
\centering
\includegraphics[width=1.0\linewidth]{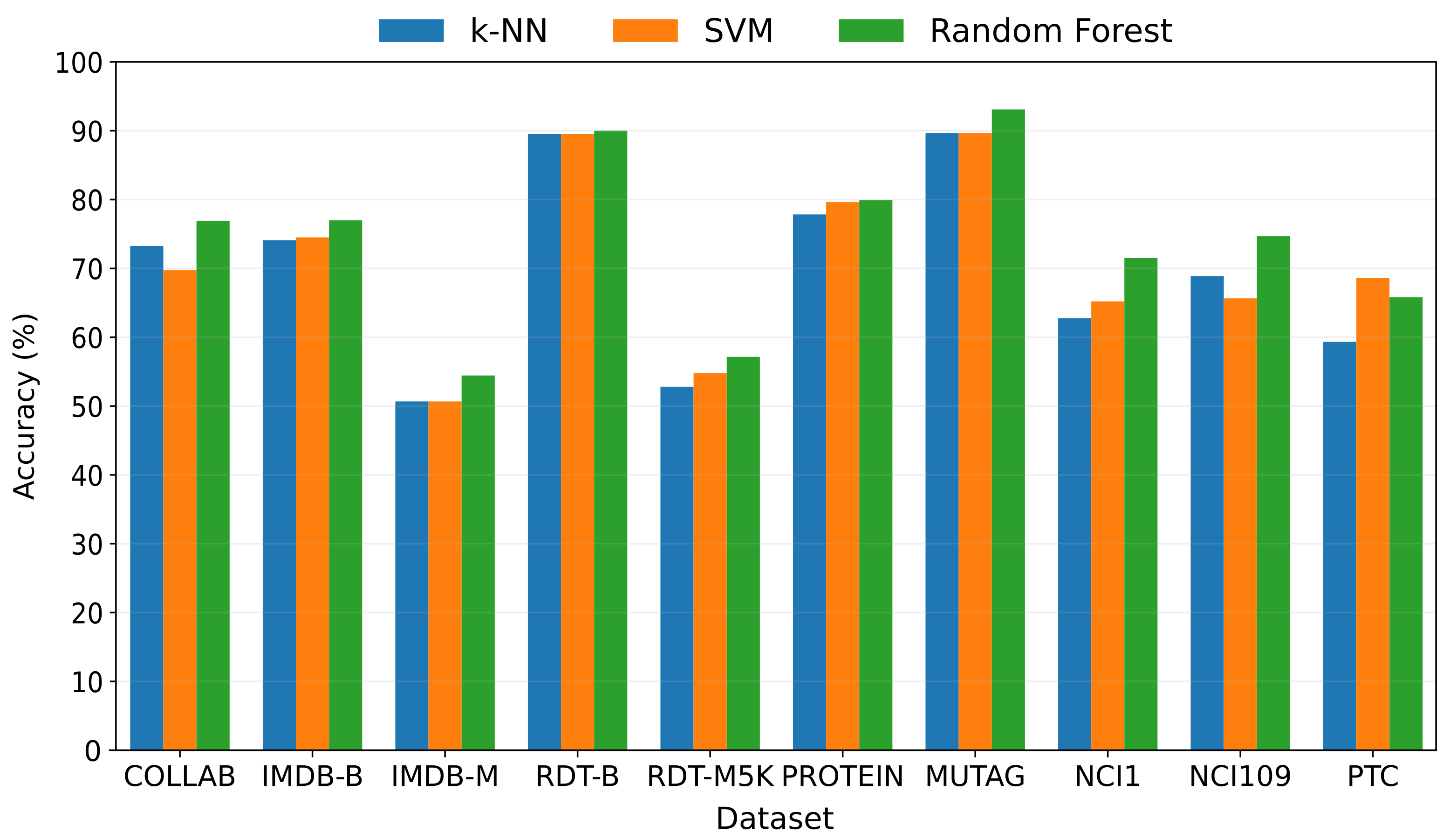}
\caption{Classification results by all the methods on all the datasets. The height of the bar indicates the accuracy percentage for the corresponding method on the corresponding dataset.}
\label{fig:all_results}
\end{figure*}

We conducted a five-fold cross-validation for each dataset and shows the accuracy results obtained by the k-NN, SVM, and Random Forest in Fig.~\ref{fig:all_results}. The height of each bar in Fig.~\ref{fig:all_results} represents the percentage of accuracy of the cross validation for each classification method. 

\begin{figure*}
\centering
\includegraphics[width=0.75\linewidth]{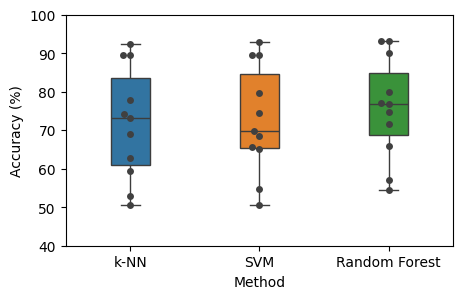}
\caption{Consistency of the performances of all methods across all datasets. Each dot represents the accuracy percentage achieved by the corresponding method on a dataset. The horizontal line inside the box indicates the mean value of all performances produced by the corresponding method.}
\label{fig:perform_consist}
\end{figure*}

We found that the classification results achieved from the ten datasets are consistent across the methods. Although, we obtained slightly better results with the Random Forest method, the classification accuracy remains almost the same regardless of the classifiers we use. To further substantiate the claim of the consistency of the performance, we conducted an ANOVA test on the performance of these three classifiers across ten datasets. The result of this test revealed no significant difference in accuracy percentages among the methods (p-value: 0.7961728 which supports the null hypothesis that there is no significant difference among the performances). Moreover, we highlighted the high correlation values between the performances of the classifiers: 0.97 between k-NN and SVM, 0.99 between k-NN and Random Forest, and 0.97 between SVM and Random Forest. These high correlation values indicate that the methods produce similar results. Furthermore, we presented the distribution of the results in a violin plot (Fig.~\ref{fig:perform_consist}) to visually illustrate the consistency of the classifiers' performance.

\begin{figure*}[htp]
\centering
\includegraphics[width=1.05\linewidth]{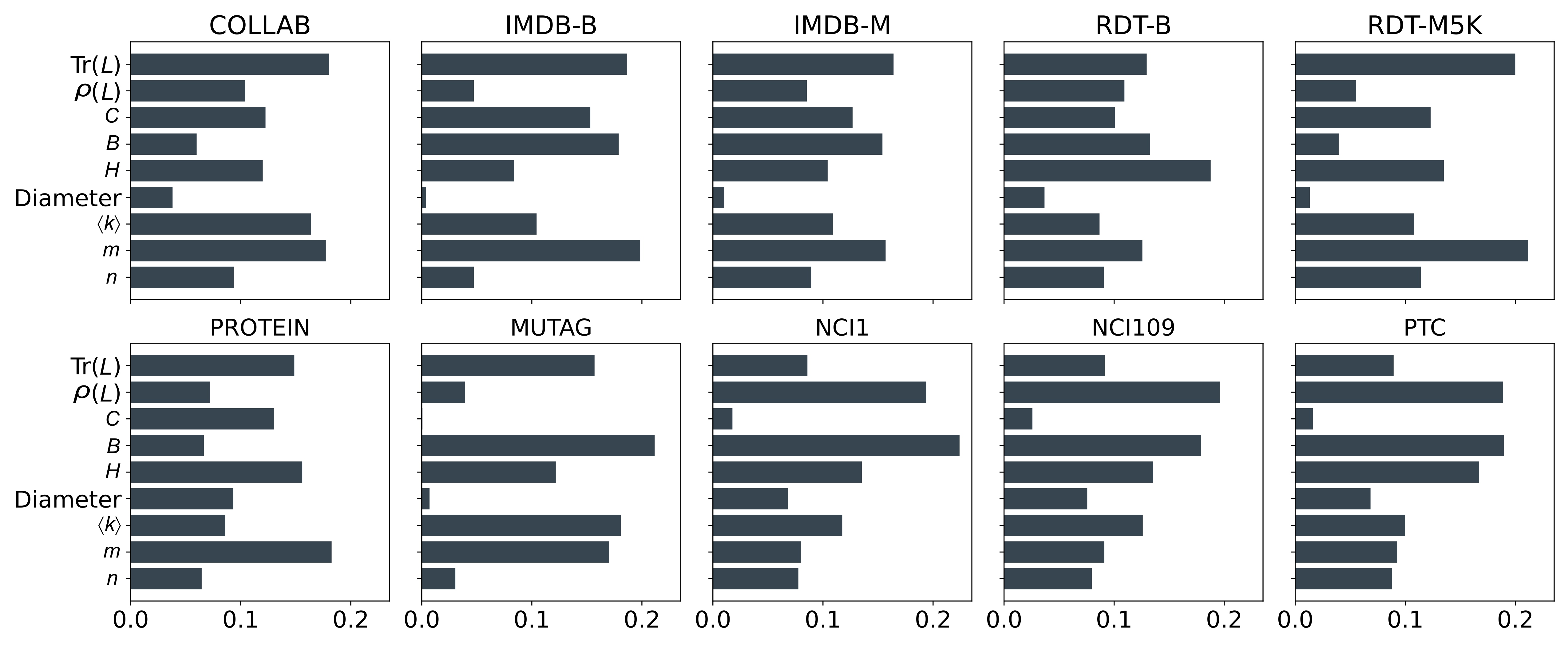}
\caption{The importance of the graph features for the Random Forest method across ten datasets. The length of the bars indicate the importance of the nine graph features: number of nodes ($n$), number of edges ($m$), average node degree ($\langle k \rangle$), average closeness centrality ($H$), average betweenness centrality ($B$), average clustering coefficient ($C$), the spectral radius ($\rho(L)$), and trace of the Laplacian matrix ($\text{tr}(L)$) in distinguishing different graph classes.}
\label{fig:feature_importance}
\end{figure*}

We also computed feature importance for the Random Forest method across ten datasets to identify the features most effective at classifying graphs into different categories. The importance of various graph features are shown in Fig.~\ref{fig:feature_importance}. Notably, no singular winning feature emerged consistently across all datasets. However, we found common top-ranking features across the datasets, which includes the number of edges ($m$), average node degree ($\langle k \rangle$), average betweenness centrality ($B$), the spectral radius ($\rho(L)$), and the trace of the Laplacian matrix ($\text{Tr}(L)$).

Moreover, we compare the results obtained by using our approach with numerous existing benchmark graph learning methods. In Table~\ref{tab:comparison-soci}, and~\ref{tab:comparison-bio}, we show the results achieved by different machine learning algorithms using the five social and five biological networks. The methods include Graphlet Kernel (GK)~\cite{shervashidze2009efficient}, Weisfeiler-Lehman Kernel (WL)~\cite{shervashidze2011weisfeiler}, Graph Convolutional Network (GCN)~\cite{kipf2016semi}, Graph Attention Network (GAT)~\cite{velivckovic2017graph}, GraphSAGE~\cite{hamilton2017inductive}, Graph Isomorphism Network (GIN)~\cite{xu2018powerful}, Deep Graph Convolutional Neural Network (DGCNN)~\cite{zhang2018end},  Capsule Graph Neural Network (CapsGNN)~\cite{xinyi2018capsule}, Discriminative Structural Graph Classification (DSGC)~\cite{seo2019discriminative}, Graph Feature Network (GFN)~\cite{chen2019powerful}, and Graph Classification by Graph Structure Learning~\cite{huynh2022graph}. In Table~\ref{tab:comparison-soci}, and~\ref{tab:comparison-bio}, the bold faces in each column  represent the top three performances on the dataset in that column. Also, we remark the best result by $^{\blacktriangle}$ and the second best result by $^{\blacklozenge}$, and the third best result by $^{\blacktriangledown}$. In our approach, the accuracy achieved by all three classifiers outperforms the best method results on PROTEINS dataset. Also, the Random Forest bets state-of-the-art performance on MUTAG, IMDB-B, and IMDB-M datasets. The other two classifiers achieved very competitive accuracy with the best results of the benchmark methods. Though the performance of SVM on the COLLAB dataset is not reasonably good compared to the best one, the k-NN and Random Forest obtained the accuracy which is very close to the leading results of the ten benchmark methods.

To assess the superior or competitive performance of our approach relative to state-of-the-art methods, we have compiled a summary table (Table~\ref{tab:top3_finishes}) that highlights the top-3 rankings across ten datasets. In Table~\ref{tab:top3_finishes}, the row labeled "Rank 1" indicates the number of times each method achieved the best performance across ten datasets. The rows "Rank 2" and "Rank 3" represent the frequency with which each method secured the second and third positions in terms of accuracy percentage, respectively. The final row aggregates the total number of top-three finishes—whether first, second, or third—for each method across all datasets. The table showed that the Random Forest attained the highest performance in four instances, whereas SVM achieved this in only one dataset. Additionally, Random Forest secured the second position once and the third position twice. In contrast, SVM obtained second and third positions only once each, and k-NN achieved third place in two instances. This summary underscores the competitive edge of Random Forest and provides a clear comparative overview of the performance of each method across the evaluated datasets.

\begin{table*}
\centering
\caption{\textbf{Comparison of the classification results for the social networks.} The bold faces in each column are the top three performances. $^{\blacktriangle}$ : indicates the best performance, $^\blacklozenge$: indicates the second best performance, and $^{\blacktriangledown}$: indicates the third best performance.}
\label{tab:comparison-soci}
\begin{tabular}{llllll}
              ~ & COLLAB & IMDB-B & IMDB-M & RDT-B & RDT-M5K \\ \midrule
        GK~\cite{shervashidze2009efficient} &  72.84 & 65.87 & 43.89 &78.00  & 41.01\\
        WL~\cite{shervashidze2011weisfeiler}  & 79.02 & 73.4 & 49.33 & 81.10 & 49.44 \\
        GCN~\cite{kipf2016semi} &  \textbf{81.72}$^\blacktriangle$ & 73.30 & 51.20 & 89.30 &  56.81 \\
        GAT~\cite{velivckovic2017graph} & 75.80 & 70.50 & 47.80 & N/A &   N/A \\
        GraphSAGE~\cite{hamilton2017inductive}  & 79.70 & 72.40 & 49.90 &  86.10 & 50.00  \\
        GIN~\cite{xu2018powerful} &  \textbf{80.20}$^{\blacktriangledown}$  & \textbf{75.10}$^\blacklozenge$ & \textbf{52.30}$^\blacklozenge$  & \textbf{92.40}$^\blacktriangle$ & \textbf{57.50}$^\blacklozenge$\\
        DGCNN~\cite{zhang2018end}&   73.76 & 70.03 & 47.83 & 77.10 & 48.70  \\
        CapsGNN~\cite{xinyi2018capsule}& 79.62 & 73.10 & 50.27 & N/A & 52.88 \\
        DSGC~\cite{seo2019discriminative} & 79.20 & 73.20 & 48.50 & \textbf{92.20}$^\blacklozenge$ & N/A  \\
        GFN~\cite{chen2019powerful}&  \textbf{81.50}$^\blacklozenge$ & 73.00 & \textbf{51.80}$^{\blacktriangledown}$  & N/A & \textbf{57.59}$^\blacktriangle$  \\
        GC-GSL~\cite{huynh2022graph} & N/A &68.46 & 46.39  & N/A & N/A \\
        k-NN & 73.25 & 74.10& 50.67 & 89.50& 52.80 \\
        SVM & 69.76 & \textbf{74.50}$^{\blacktriangledown}$  & 50.67 & 89.53& 54.80 \\
        Random Forest  & 76.90 & \textbf{77.00}$^\blacktriangle$ & \textbf{54.44 }$^\blacktriangle$ & \textbf{90.00}$^{\blacktriangledown}$  &  \textbf{57.15}$^{\blacktriangledown}$ \\
\bottomrule
\end{tabular}
\end{table*}

\begin{table*}
\centering
\caption{\textbf{Comparison of the classification results for the biological networks.} The bold faces in each column are the top three performances. $^{\blacktriangle}$ : indicates the best performance, $^\blacklozenge$: indicates the second best performance, and $^{\blacktriangledown}$: indicates the third best performance.}
\label{tab:comparison-bio}
\begin{tabular}{llllll}
              ~ & PROTEINS & MUTAG &   NCI1 & NCI109 & PTC\\
               \midrule
        GK~\cite{shervashidze2009efficient} & 71.67 & 81.58 & 69.18 & 69.82 & 55.65\\
        WL~\cite{shervashidze2011weisfeiler}  & 74.68 & 82.05 & 80.13 & \textbf{80.22}$^{\blacktriangledown}$ &  57.97\\
        GCN~\cite{kipf2016semi} & 75.65 & 87.20&  \textbf{83.63}$^\blacktriangle$   &  72.50  & N/A\\
        GAT~\cite{velivckovic2017graph} & 74.70& 89.40 & 74.90 & 75.80 & N/A \\
        GraphSAGE~\cite{hamilton2017inductive} & 65.90 & 79.80 &  76.0 & 70.30 & 63.90\\
        GIN~\cite{xu2018powerful} &  76.20 & 89.40 & \textbf{82.70}$^{\blacktriangledown}$ & \textbf{82.00}$^\blacktriangle$  & \textbf{64.60}$^{\blacktriangledown}$\\
        DGCNN~\cite{zhang2018end} & 75.54 & 85.83 &  74.44 & 75.03 & 58.59\\
        CapsGNN~\cite{xinyi2018capsule} &  76.28 & 86.67 & 78.35 & N/A & N/A\\
        DSGC~\cite{seo2019discriminative} & 74.20 & 86.70 &  79.80 & N/A & N/A\\
        GFN~\cite{chen2019powerful}& 76.46 & \textbf{90.84}$^\blacklozenge$  &  \textbf{82.77}$^\blacklozenge$  & N/A & N/A \\
        GC-GSL~\cite{huynh2022graph} & 76.55 & 83.86 & 82.04 & \textbf{81.86}$^\blacklozenge$  & 60.11\\
        k-NN & \textbf{77.84}$^{\blacktriangledown}$ & \textbf{89.66}$^{\blacktriangledown}$ & 62.77 & 68.89 &  59.35 \\
        SVM &  \textbf{79.64}$^\blacklozenge$ & \textbf{89.66}$^{\blacktriangledown}$ &  65.21 & 65.65& \textbf{68.61}$^\blacktriangle$ \\
        Random Forest  & \textbf{79.93}$^\blacktriangle$  & \textbf{93.10}$^\blacktriangle$ &  71.53 &74.68 & \textbf{65.81}$^\blacklozenge$ \\
\bottomrule
\end{tabular}
\end{table*}

\begin{table}
\centering
\tiny{
\begin{tabular}{lllllllllllllll}
\toprule
Method     & k-NN & SVM & Random & GK & WL & GCN & GAT & Graph- & GIN & DGCNN & CapsGNN & DSGC& GFN & GC-GSL\\
& & & Forest & & & &  & SAGE& & & & & &\\
\midrule
Rank 1   & 0             & 1            & 4                     & 0           & 0           & 2            & 0            & 0                & 2            & 0              & 0               & 0            & 1           & 0              \\
Rank 2   & 0             & 1            & 1                     & 0           & 0           & 0            & 0            & 0                & 3            & 0              & 0               & 1            & 3           & 1              \\
Rank 3   & 2             & 1            & 2                     & 0           & 1           & 0            & 0            & 0                & 3            & 0              & 0               & 0            & 1           & 0              \\
\hline
Total & 2 & 3 & 7 & 0 & 1 & 2 & 0 & 0 & 8 & 0 & 0 &1 & 5 & 1\\
\bottomrule
\end{tabular}}
\caption{Summary of top-3 performances across all datasets. The second row lists the number of times each method achieved the highest performance. The third and fourth rows denote the frequency of each method securing the second and third positions, respectively. The final row aggregates the total number of top-three finishes (first, second, or third) for each method in the classification performance across all datasets.}
\label{tab:top3_finishes}
\end{table}

\begin{figure}
\centering
\includegraphics[width=.85\linewidth]{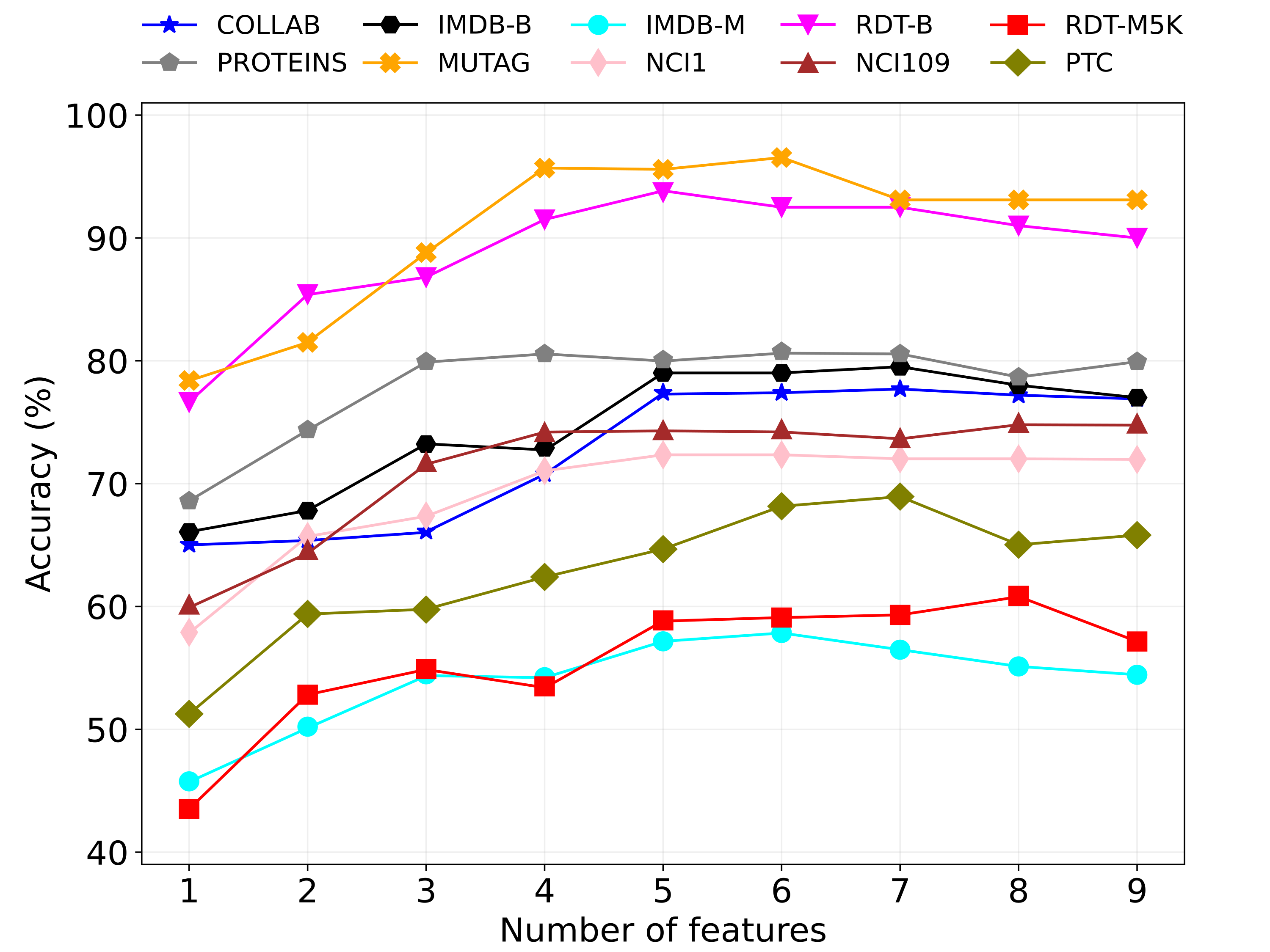}
\caption{Efficacy of feature combinations in graph classification. The x-axis represent the number of features included in the respective best combination. The y-axis represents the accuracy percentage of the classification results obtained by Random Forest on the best feature combinations.}
\label{fig:feature_models}
\end{figure}

Furthermore, we systematically investigate the efficacy of feature combinations in graph classification, spanning from one to nine features. Beginning with single feature model, we iteratively evaluate the performances on each single feature and then select the best single feature. Subsequently, we explore two feature model by considering all ${9 \choose 2} = 36$  possible feature combinations and identify the best combination. Similarly, we select the best combination consisting three features by examining all ${9 \choose 3} = 84$ feature combinations  for the three feature model. Analogously, we continue this process for the $k=4,5, \cdots, 8$ feature models to find the best combinations of $k$ features by assessing the results from $9 \choose k$ feature combinations. For nine feature model we consider the results that reported in the Fig.~\ref{fig:all_results}. Throughout this process, we exclusively employed only the Random Forest model for the classification. We present the classification results on the best features combination for $k = 1,2,\cdots, 9$ in Fig.~\ref{fig:feature_models}. Surprisingly, across the majority datasets, we see that the classification accuracy achieved with fewer features than nine features suppresses that obtained with the full set of nine features. However, it is noticeable that these results are obtained by an exhaustive and time-intensive exploration of the best feature combinations. Additionally, it is crucial to note that the best feature combinations for $k=1, 2, \cdots, 8$ are not identical across the datasets. This variability of features in the best set underscores the nuanced nature of feature selection and its dependence on the specific attributes of each dataset.

\section{Discussion}\label{sec:discussion}
In this study, we propose a general machine learning approach for graph classification that is systematic, simple yet effective. The approach is based on a feature vector derived from nine graph structural properties. These properties encapsulate discriminating structural information, enabling the accurate classification of different graph classes. Notably, our method surpasses state-of-the-art machine learning techniques on certain datasets and achieves competitive accuracy across ten benchmark datasets. Among the three classifiers employed—k-Nearest Neighbors (k-NN), Support Vector Machine (SVM), and Random Forest—the Random Forest classifier demonstrates superior performance. However, the consistency of results across the different classifiers underscores the robustness of our approach. This robustness suggests that any standard classification method can effectively classify graphs with high accuracy using the features proposed in this study. In the feature importance analysis for the Random Forest method, we found that no consistent single feature distinguishing graph classes. However, common top-ranking features across various datasets include number of edges, average node degree, average betweenness centrality, and trace of the Laplacian matrix. Interestingly, our analysis of the best feature combinations reveals that in many cases, using fewer than nine features can yield higher classification accuracy than employing the full set. This finding highlights the potential for further optimization of feature selection, though it was achieved through a rigorous and time-intensive exploration of optimal feature combinations.

Furthermore, this study opens several avenues for future work. First, the methodology presented here is easily adaptable to node classification, which is essential for understanding network behavior at a finer granularity. Additionally, the study aims to identify the least number of essential graph properties that effectively capture discriminative structural information, contributing to a more streamlined and interpretable classification process. While the graphs considered in this study are modest in size and suitable for classification studies, our methodology is capable to accommodate larger graphs. However, we aim to enhance the scalability of our methodology for large graphs by incorporating subgraph sampling methods to reduce computational demands. In the comparison, our approach, which relies solely on nine graph structural features, consistently demonstrated either superior or competitive performance when measured with the state-of-the-art GNN methods that utilize node attributes when available. While our study focuses exclusively on structural properties, future work will extend our research to datasets that includes node and edge attributes. By fully leveraging GNNs, we aim to provide a more comprehensive comparison and further enhance classification accuracy. Moreover, we envision applying their classification approach across a wider range of fields beyond those covered by the benchmark datasets, further enhancing the applicability and impact of our methodology.  Overall, this research enhanced the efficiency, interpretability, and applicability of graph classification methodologies, opening new possibilities for their utilization in diverse scientific and practical contexts. Our approach not only contributes to the current understanding of graph classification but also provides a robust framework for future advancements in the field.

\section*{Credit authorship contribution statement} Saiful Islam conceived, and designed  the study.  Saiful Islam and Md. Nahid Hasan collected the data. Saiful Islam, Md. Nahid Hasan, and Pitambar Khanra contributed to the analysis. Saiful Islam accomplished the visualization and Saiful Islam, Pitambar Khanra validated the outcomes. Saiful Islam, Md. Nahid Hasan, and Pitambar Khanra wrote and edited the manuscript. All the authors read and approved the final version of the manuscript. 

\section*{Declaration of competing interest} The authors declare that they have no known competing financial interests or personal relationships that could have appeared to influence the work reported in this paper.

\section*{Acknowledgement}
We thank Prof. Naoki Masuda for his invaluable insights and suggestions, which have enhanced the quality and depth of this work.

\section*{Declaration of generative AI and AI-assisted technologies in the writing process}
During the preparation of this work the authors used language refinement tools in order to refine the english style of the presentation alongside the grammatical corrections. After using this tool/service, the authors reviewed and edited the content as needed and takes full responsibility for the content of the publication.

\bibliography{bibliography}
\end{document}